\documentclass[journal]{IEEEtran}

\usepackage{graphicx}
\pdfoutput=1
\usepackage{epstopdf}
\usepackage[numbers,sort&compress]{natbib}

\usepackage{caption3}
\usepackage{amssymb}
\usepackage{amsmath}

\usepackage{amsfonts}

\usepackage{bbding}
\usepackage{array,diagbox,siunitx}

\usepackage{cases}
\usepackage{float}
\usepackage{multirow}
\usepackage{mathrsfs}
\usepackage{algorithmic}
\usepackage[ruled]{algorithm2e}
\usepackage{rotating}
\usepackage{balance}
\usepackage{threeparttable}
\usepackage{color}
\usepackage{colortbl}
\definecolor{mygray}{gray}{.9}
\usepackage{makecell}
\usepackage{pdflscape}
\usepackage{siunitx}
\usepackage{stfloats}
\usepackage{verbatim}
\usepackage{setspace}
\usepackage{threeparttable}
\usepackage{supertabular,booktabs}
\usepackage{rotating}
\usepackage{supertabular}
\usepackage{pifont}
\usepackage{xcolor,pict2e}

\usepackage{tikz}
\usepackage{booktabs}
\usepackage{footnote}
\usepackage[pagewise]{lineno}
\usepackage{balance}

\usepackage{hyperref}
\usepackage{ulem}

\newcommand{\etal}{\textit{et al.}}
\newcommand{\ie}{\textit{i.e.}}
\newcommand{\eg}{\textit{e.g.}}

\begin{document}
\title{Incorporating Scene Context and Semantic Labels for Enhanced Group-level Emotion Recognition}

\author{Qing Zhu,
        Wangdong Guo,
        Qirong Mao,~\IEEEmembership{Member,~IEEE},
        Xiaohua Huang,~\IEEEmembership{Senior Member,~IEEE,}
        Xiuyan Shao,
        Wenming Zheng,~\IEEEmembership{Senior Member,~IEEE}
\thanks{{Q. Zhu, W. Guo and Q. Mao are with the School of Computer Science and Communication Engineering, Jiangsu University, Zhenjiang 212013, China. (E-mail: zhuqing@stmail.ujs.edu.cn, 2212308045@stmail.ujs.edu.cn, mao\_qr@ujs.edu.cn)}}

\thanks{X. Huang with the Oulu School, Nanjing Institute of Technology, Nanjing 211167, China.
(E-mail: xiaohuahwang@gmail.com)}

\thanks{X. Shao is with the School of Management, Southeast University, Nanjing, Jiangsu, China (e-mail: xiuyan\_shao@seu.edu.cn)}

\thanks{W. Zheng is with the Key Laboratory of Child Development and Learning Science (Southeast University), Ministry of Education, Southeast University, Nanjing 210096, China.
(E-mail: wenming\_zheng@seu.edu.cn)}
}


%
\maketitle
\begin{abstract}
Group-level emotion recognition (GER) aims to identify holistic emotions within a scene involving multiple individuals. Current existed methods underestimate the importance of visual scene contextual information in modeling individual relationships. Furthermore, they overlook the crucial role of semantic information from emotional labels for complete understanding of emotions. To address this limitation, we propose a novel framework that incorporates visual scene context and label-guided semantic information to improve GER performance. It involves the visual context encoding module that leverages multi-scale scene information to diversely encode individual relationships. Complementarily, the emotion semantic encoding module utilizes group-level emotion labels to prompt a large language model to generate nuanced emotion lexicons. These lexicons, in conjunction with the emotion labels, are then subsequently refined into comprehensive semantic representations through the utilization of a structured emotion tree. Finally, similarity-aware interaction is proposed to align and integrate visual and semantic information, thereby generating enhanced group-level emotion representations and subsequently improving the performance of GER. Experiments on three widely adopted GER datasets demonstrate that our proposed method achieves competitive performance compared to state-of-the-art methods. 

\end{abstract}

\IEEEpeerreviewmaketitle
\section{Introduction}
\label{sec:intro}

\begin{figure}
    \centering
    \includegraphics[width=1.0\linewidth]{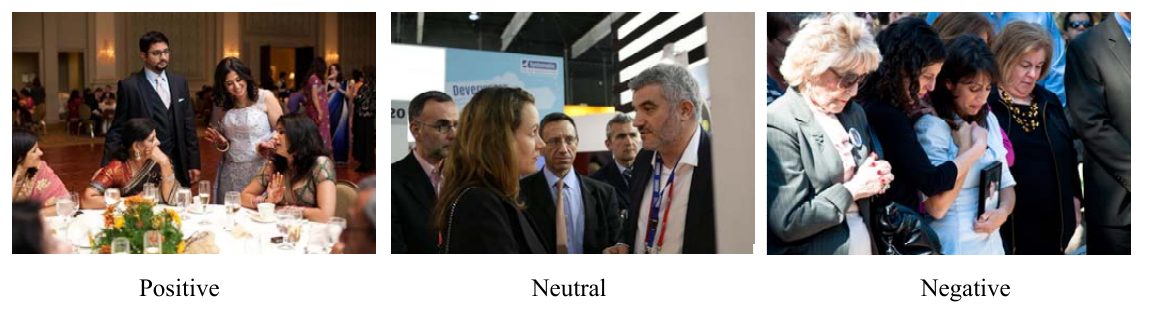}
    \caption{Examples corresponding to the three categories from the GER dataset.}
    \label{fig:sample}
    \vspace{-10pt}
\end{figure}

\IEEEPARstart {A}UTOMATIC recognition of human emotions, a well-explored area within multimedia computing that includes image, audio, text, and video analysis, has significantly advanced our understanding of human behavior \cite{TurchetOOG24,SongGYTW24,johnson2021exploring}. In recent decades, substantial progress has been made in recognizing emotions at the individual level \cite{YinJHYW24,RodriguezFP25,MaoZZSH22}. Social behavior research \cite{barsade2015group,barsade2012group} indicates that people often change their reactions and behaviors based on their perceptions of others' emotions. As a result, there has been increased interest in group-level emotion recognition (GER). GER seeks to classify the collective emotion of multiple individuals into three categories: positive, neutral, and negative (as illustrated in Fig.~\ref{fig:sample}), which has significant societal implications. It is relevant to various fields such as social behavior analysis, public security, and human-robot interactions~\cite{SanchezHTH20,VeltmeijerGH23}. 

\begin{figure}
    \centering
    \includegraphics[width=1.0\linewidth]{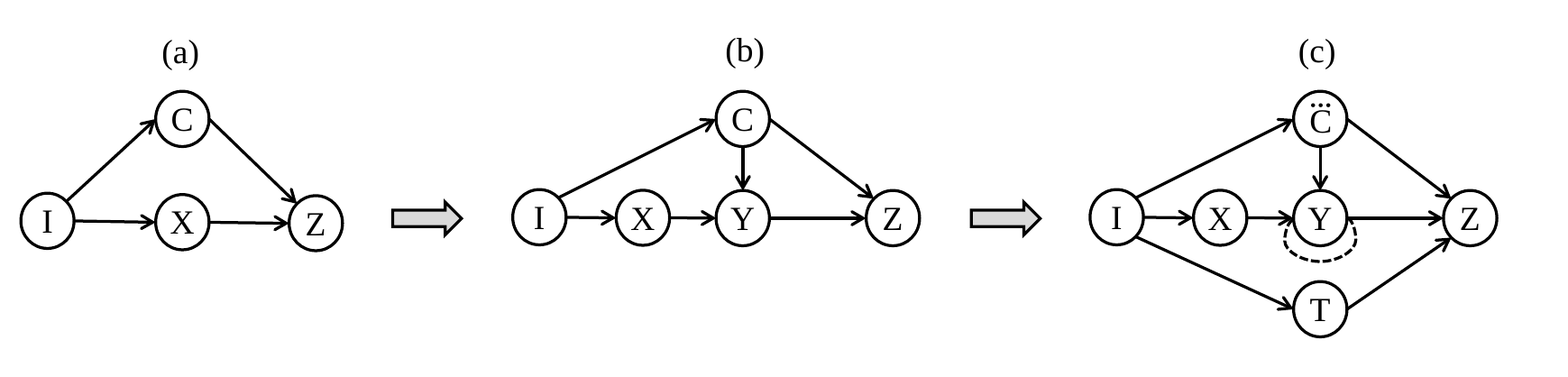}
    \caption{(a) Causal graph of early models. (b) Causal graph of existing context-based models. (c) Causal graph of our proposed model. I: input data. C: global scene feature. X: individual features. Y: context-encoded individual features. T: semantic embeddings. Z: group-level emotion prediction. }
    \label{fig:Schematic plot}
        \vspace{-10pt}
\end{figure}

Given a group-level emotion image, the essence of GER is in modeling the collaborative relationships among individuals and comprehending the contextual information within the scene. Previous works \cite{huang2019analyzing,KhanLCMOT18,surace2017emotion} have employed deep learning techniques to explore diverse emotion-related features and effectively aggregate them for group-level emotion inference. We revisit these methods through the lens of a causal graph, as illustrated in Fig. \ref{fig:Schematic plot}(a). In this graph, $I \rightarrow C$ represents the generation of global scene features from the given image, $I \rightarrow X$ signifies the extraction of individual features (\eg, face, object)  with the help of corresponding detectors from the input image, $C \rightarrow Z$ and $X \rightarrow Z$ depict the integration of scene features and aggregated individual features to generate group-level representations for emotion classification, respectively. However, these methods primarily focus on localizing individual features while neglecting the broader global scene context, which is crucial to enhancing GER.

To resolve the above issue, recent works \cite{fujii2020hierarchical,Khan0CT21,zhang2022semi,WangZTL23} have leveraged the global scene context to facilitate interactions among individuals,  as illustrated by the corresponding causal graph in Fig. \ref{fig:Schematic plot}(b). The primary distinctions are $C \rightarrow Y$: generates the context-encoded individual features through weighted fusion with the global scene feature and primordial individual features, and $Y \rightarrow Z$: the context-encoded individual features are then utilized to derive context-aware representations, rather than reasoning directly on the individual features. Several researchers~\cite{fujii2020hierarchical,Khan0CT21}  straightforwardly concatenated the scene information with individual characteristics and then used attention mechanisms to generate fusion weights, while achieving notable results. Furthermore, Wang~\etal ~\cite{WangZTL23} employed cosine similarity between the global scene and individual features in the subspace to determine the fusion weights, achieving state-of-the-art performance in recognition accuracy. The shortcomings of these methods are reflected in two aspects: (1) treating scene information and individuals on a one-to-one basis, which weakens the encoding of global scene context into individual interaction modeling; (2) struggling to capture comprehensive context from complex scenes due to insufficient semantic information.

In fact, given that all individuals coexist within the same scene, it is crucial to model interactions among individuals, guided by the global scene context. However, existing methods primarily rely on visual modality alone, neglecting the potential benefits of incorporating semantic-level information associated with emotion labels. This limitation hinders the advancement of GER. Effective GER requires not only an understanding of the visual scene but also the integration of explicit semantic guidance to refine emotional interpretation. For example, when identifying a “positive” image, as illustrated in Fig.~\ref{fig:sample}, it is imperative to analyze individuals, their interactions, and the surrounding environment holistically. The semantic information linked to the “positive” label provides valuable contextual cues, such as smiles, expressive body language, traditional attire, and festive decorations. These factors collectively enhance the perception of conveyed emotions, thereby improving the accuracy of group-level emotion inference.

To address the aforementioned drawbacks, we propose a novel method guided by a newly designed causal graph, as illustrated in Fig.~\ref{fig:Schematic plot}(c). A key distinction of our method from existing context-based approaches is its ability to incorporate multi-scale scene context information $\overset{\cdots}{C}$ to enhance the encoding of individual features and effectively model their relationships $Y \rightarrow Y$. Furthermore, we introduce a novel branch $I \rightarrow T$ that leverages image labels to generate emotion semantic embeddings, thereby refining the overall emotion representation $Z$.

In practice, we propose a novel reasoning paradigm that integrates visual scene context with label-guided semantic information to enhance GER. Unlike previous methods that rely heavily on the entire complex scene, our approach utilizes multi-scale scene features to encode interactions among individual features in a diversified manner, leveraging a cross-attention mechanism. Specifically, we align individual features $X$ with global scene features $C$ across multiple scales, aggregating and refining contextual information to generate diverse context-encoded representations $Y$. This process effectively mitigates the ambiguity arising from scene complexity. Furthermore, we derive label-guided semantic embeddings $T$ to enhance group-level emotion representation. To enrich semantic understanding beyond simple class labels, we employ a Large Language Model (LLM) such as ChatGPT, to generate an extensive collection of nuanced emotion lexicons pertinent to group attribute. Subsequently, we utilize a Graph Convolutional Network (GCN) to learn discriminative semantic representations by traversing a  structured emotion tree, effectively linking emotion lexicons to their corresponding emotion labels. Finally, we align and integrate visual and semantic feature to contruct a more robust representation of group-level emotions for emotion prediction $Z$. Experimental results demonstrate that our proposed method outperforms state-of-the-art methods on three widely adopted GER datasets. 

The main contributions of this paper are summarized as follows:
\begin{itemize}
    \item We propose a novel reasoning paradigm that effectively integrates rich visual scene context and label-related semantics within a new causal graph framework, meticulously crafted to enhance GER.

    \item It is the first time in GER that label-related semantics have been embedded, overcoming the limitations of conventional approaches that relied purely on visual features. By prompting a LLM with group-level emotion labels to generate nuanced emotion lexicons, a comprehensive semantic understanding is facilitated, thereby enhancing the discriminability of predicted group-level emotion representations.
    
    \item Extensive evaluations on three public GER datasets, \ie, GAFF2, GAFF3, and GroupEmoW, demonstrate that our approach achieves competitive performance compared to existing methods.
\end{itemize}

\section{Related Work} 
\subsection{Group-level Emotion Recognition}
In recent years, GER has received increasing attention, leading to the development of numerous advanced approaches. Much of the research in this domain focuses on integrating diverse emotional cues to infer group-level emotions~\cite{guo2020graph, GuoZPBB18, KhanLCMOT18, Khan0CT21, quach2022non,WangZTL23}. As GER research evolves, recent studies have emphasized the significance of incorporating supplementary contextual factors, such as objects and scenes, to improve recognition accuracy. Fujii~\etal~\cite{fujii2020hierarchical} proposed a hierarchical model that first classifies facial features and subsequently merges object and scene information. Similarly, Dai~\etal~\cite{DaiLSL19} employed Long Short Term Memory (LSTM) networks to model individual features, enhancing group emotion representation. These advances show the pivotal role of context integration in improving GER accuracy and robustness.

The rapid progress in deep learning have facilitated methods that encode global context into individual features while refining aggregated information. Several studies have utilized Graph Neural Networks (GNNs) to model interactions among contextually enriched individual features, leading to improved group-level emotion understanding. For instance, Guo~\etal~\cite{guo2020graph} utilized GNNs to model relationships among face, object, and skeleton features based on holistic scene comprehension. More recently, attention-based methods have been employed to enhance the modeling of individual interactions within a scene, yielding significant improvements in GER.
Khan~\etal~\cite{Khan0CT21} introduced regional attention and context-aware fusion to assign appropriate weights to emotion streams. Wang~\etal~\cite{WangZTL23} mapped scene and individual features into a shared subspace for similarity-based fusion. Additionally, Xie~\etal~\cite{XieLCCLSC23} designed cross-patch attention within vision transformers to integrate scene contexts and facial regions effectively.

Despite these advances, existing methods primarily integrate global scene and individual features in a one-to-one manner, limiting the effectiveness of global context efficacy and neglecting semantic emotion cues. This constraint hinders a comprehensive understanding of group emotions. Effective GER requires a synergistic approach that models both visual contexts and emotion-guided semantics. Our approach addresses these limitations by jointly leveraging multi-scale scene contexts and label-aligned emotion semantics, thereby enhancing GER performance.

\subsection{Semantic encoding based on labels}
Emotion label information plays a pivotal role in emotion analysis tasks by providing semantic guidance, enriching emotion embeddings, and facilitating multi-modal feature fusion~\cite{ZhangCSW22, JiangLZL23}. Labels enable models capture features relevant to each emotion category, enhance the alignment of multi-modal data, and serve as a foundation for constructing loss functions or regularization terms~\cite{LiNZ023, ChenHHK23, DengR23a}. Moreover, they contribute to a more granular recognition of emotions and support the construction of emotion knowledge graphs, which capture relationships among various emotions, ultimately leading to more accurate and context-aware predictions~\cite{PengWKNLC22, WangLSLTF22}.

The semantic encoding of labels has been widely applied in various image and video understanding tasks, including video caption~\cite{SongGYTYW23}, image classification~\cite{ChenGZSQDL22, LiuWJCL24, PengYXWX24}, facial expression recognition~\cite{ZhaoP23} and group activity recognition~\cite{LiuZ021, WuTXGS24}. For instance, Liu~\etal~\cite{LiuZ021} directly utilized group activity labels to construct a semantic graph, refining visual representations. Song~\etal~\cite{SongGYTYW23} introduced an emotion-prior awareness network that integrates catalog-level psychological categories with lexical-level common words to achieve precise and detailed emotion recognition. More recently, the use of LLMs for obtain semantic information to support research tasks has seen rapid development~\cite{ChengWTZ25,JungSCR024}. Zhao~\etal~\cite{ZhaoP23} proposed a vision-language model that leverages expression descriptors related to facial behavior, generated by LLMs based on emotion labels, as textual input to enhance expression understanding.

\begin{figure*}
    \centering
    \includegraphics[width=0.9\linewidth]{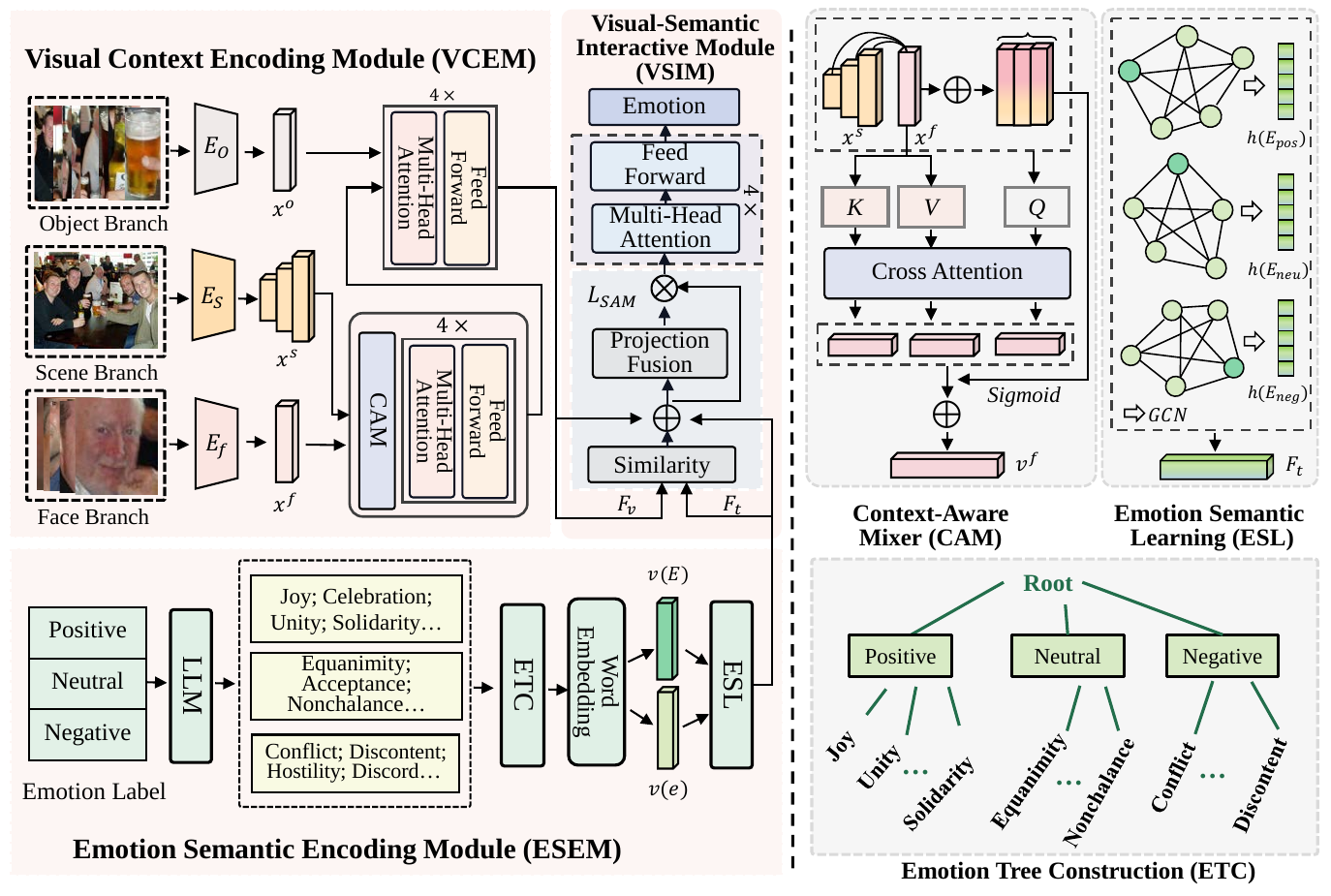}
    \caption{Framework of our proposed method. The framework comprises three key modules: (1) a visual context encoding module, (2) an emotion semantic encoding module, and (3) a visual-semantic interactive module. First, the visual context encoding module extracts emotion-related features from face, scene context, and object cues using CNN-based backbones. The extracted facial features are refined through a context-aware mixer, which integrates multi-scale scene features. These refined facial features are then combined with object features to form comprehensive visual embeddings. Next, the emotion semantic encoidng module utilizes LLM to generate nuanced emotion lexicons guided by emotion labels. Both the lexicons and labels are encoded into semantic representations and structured into an emotion tree. GCNs then process and refine these structured representations, producing enriched semantic representations. Finally, the visual-semantic interactive module aligns and fuses the visual and semantic representations, generating refined representations for emotion prediction.}
    \label{fig:framework}
    \vspace{-10pt}
\end{figure*}

\section{Proposed Method}
The proposed framework (Fig.~\ref{fig:framework}) consists of three key modules: Visual Context Encoding (VCEM), Emotion Semantic Encoding (ESEM), and Visual-Semantic Interaction (VSIM). Given an input image, the framework first detects facial and object regions and extracts their features extracted using CNN-based encoders. To enhance contextual understanding, scene and face features are refined through a Context-Aware Mixer (CAM), which models interactions between these elements. The refined facial features are then fused with object features via a transformer encoder. In parallel, ESEM encodes emotion-related semantic information. It processes emotion labels and LLM-generated lexicons, converting them into structured semantic embeddings. These embeddings are organized as an emotion tree and further refined using GCN-based Emotion Semantic Learning (ESL) to enhance representation quality. Finally, VSIM integrates the learned visual-semantic features through Similarity-aware Feature Fusion (SFF).  The alignment of these representations is optimized through a Similarity Alignment Matching (SAM) loss, ensuring robust feature correspondence for the final emotion classification task.

\subsection{Visual Context Encoding Module}
The VCEM is designed to effectively integrate visual features relevant to emotions by combining both local individual features (encompassing faces and objects) with global context (incorporating the overall scene). It begins by encoding context-aware details to refine individual features, ensuring a more comprehensive representation. These enriched feature are then fused through a fusion mechanism, allowing the model to capture intricate relationships between individuals and their surrounding environment. This structure integration enhance the precision of GER.

\textbf{1) Visual Representation Extraction:} The input image is divided into three primary visual cues: face, object, and scene. For face features, a VGG16 network is employed to extract local features from the detected face regions using an off-the-shelf face detector~\cite{zhang2016joint}, resulting in the set $\{x_{i}^{f}\}_{i=1}^{I}$. Object features are extracted  applied VGG16 to regions proposed by an object detector~\cite{RenHG017}, yielding $\{x_{j}^{o}\}_{j=1}^{J}$. Scene features are obtained using ResNet50 with Feature Pyramid Networks (FPN)~\cite{LinDGHHB17} to capture multi-scale representations, denoted as $\{x_{k}^{s}\}_{k=1}^{K}$. These multi-scale features are aligned via RoIAlign~\cite{HeGDG17} and subsequently projected into a latent space through fully connected layers.

\textbf{2) Context-Aware Mixer:} To enhance context integration, the CAM is designed to fuse multi-scale scene context with face individual representations. Given the face features $x^{f} = [x_{1}^{f}; x_{2}^{f}; \cdots; x_{I}^{f}]$, we prepend the $k$th scale scene feature $x^{s}_k$ to the face individual features. The CAM is formulated as follows:

\begin{equation}
\begin{aligned}
    Q & = [x^{s}_k;x^{f}] W_{Q}, K=x^{f}W_{K}, V=x^{f}W_{V},\\
    h^{f}_k & = softmax \left(\frac{QK^{T}}{\sqrt{d}}\right)V,
\end{aligned}
\end{equation}
where $[;]$ denotes feature concatenation, $W_{(\cdot)}$ represents trainable weight matrices and $h^{f}_k$ denotes the context-enhanced face individual feature at the $k$th scale. Subsequently, each scale undergoes weighted fusion:

\begin{equation}
\begin{aligned}
    w_k &= \sigma(W_{\varepsilon}[x^{s}_k; x^{f}]), \\
    v^{f} &= \sum_{k=1}^{K} w_k h^{f}_k,
\end{aligned}
\end{equation}
where $w_k$ represents scale-specific fusion weights derived from the concatenation of scene and face individual features, $\sigma$ is the sigmoid activation, and $W_{(\varepsilon)}$ is a learned weight matrix. The final aggregated face individual representation $v^{f}$, encapsulates multi-scale context information.

Within the CAM, the aggregated face individual representations $v^{f}$ are concatenated with the scene features $x^{s} = [x_{1}^{s}; x_{2}^{s}; \cdots; x_{K}^{s}]$, followed by processing through multi-head attention (MHA) and a feed-forward network (FFN), ensuring cohesive integration of contextual information:
\begin{equation}
{v}^{f;s} = FFN(MHA([v^{f}; x^{s}], [v^{f}; x^{s}], [v^{f}; x^{s}])),
\end{equation}
where ${v}^{f;s}$ represents the refined fusion of scene and face features, effectively leveraging scene context to enhance individual relationship modeling and interactions.

In the final stage, the refined face-scene representation ${v}^{f;s}$ is fused with object features $x^{o}$ to produce the overall visual embedding:
\begin{equation}
F_{v} = Fusion([{v}^{f;s}; x^{o}]),
\end{equation}
where the fusion layer leverages a standard transformer encoder
(4-layer transformer blocks within multi-head attention and feed-forward operations) to derive the unified representation $F_{v}$, effectively capturing the comprehensive visual emotion representation essential for downstream GER.

\subsection{Emotion Semantic Encoding Module}
This subsection describes the construction of an emotion tree and the structured representations of emotion semantics. By prompting a LLM to generate lexicons associated with predefined emotion labels, our approach effectively captures fine-grained emotional semantics. To model the  interconnections and relationships between these lexicons and labels, we employ GCNs to jointly refine their embedding. This process ensures that multi-granularity semantic information is seamlessly integrated, resulting in a unified and comprehensive emotion semantic representation. 

\textbf{1) Emotion Tree Construction:} 
In GER, existing approaches often struggle to capture the nuanced dynamics of group emotions due to the overly broad and generalized nature of emotion labels (\eg, positive, neutral, or negative). This limitation hinders the model’s ability to accurately interpret subtle emotional cues and complex interpersonal interactions within scene contexts. To address this issue, we leverage LLM (\eg, ChatGPT) to automatically generate fine-grained emotion lexicons, eliminating the constraints of manual design. This systematic approach produces rich, context-aware lexicons that encapsulate detailed group-level emotional dynamics, thereby enhancing the model’s ability to represent and differentiate intricate emotional variations effectively. The LLM is prompted with queries such as:

\textit{Q: Please provide me with some words that convey group-level emotion of \{\textbf{class}\}.}

\textit{A: Here are some words that convey \{\textbf{class}\} group-level emotions:$\ldots$}\\where \{\textbf{class}\} is sequentially replaced with each group-level emotion class: positive, neutral, and negative. 

The generated lexicons for each group-level emotion class are used to construct an emotion tree, providing a structured and hierarchical framework that organizes emotional semantic associated with each label. Following the hierarchical emotion structure outlined in~\cite{SongGYTYW23}, the emotion tree categorizes fine-grained emotion lexicons such as “Joy”, “Unity”, and “Solidarity” under broader emotion classes like “Positive”. This hierarchical representation enhance the model's ability to capture nuanced emotional cues at multiple levels of granularity, improving its interpretability and classification performance.

\textbf{2) Emotion Semantic Learning:} 
To learn structured representation of emotion semantics and model the relationships among class-level emotions and their lexicon-based representations, we introduce a subordinate ESL mechanism. The ESL operation consolidates and refines the generated lexicons, thereby enhancing the semantic richness of each emotion class. For each emotion class $c \in \{\text{pos, neu, neg}\}$, we define a corresponding set of emotion lexicons $\{ e_{m}^{c} \}_{m=1}^{M}$, where each lexicon $e_{m}^{c}$ represents the $m$th emotion word associated with class $c$. The embeddings for each lexicon $e_{m}^{c}$, denoted as $\mathbf{v}(e_{m}^{c})$, are extracted using GloVe~\cite{PenningtonSM14}, while the emotion class embedding $\mathbf{v}(E_{c})$ is obtained in a similar manner.

To capture the relationships among emotion lexicons and their corresponding emotion class, we construct a fully connected graph $G_{c} = (V_{c}, A_{c})$, where $A_{c}$ is the adjacency matrix constructed using the cosine similarity among lexicon embeddings and the class embedding. The node set $V_{c}$ includes both the lexicon embeddings and the corresponding class label embedding:

\begin{equation}
V_{c} = \{\mathbf{v}(e_{m}^{c})\}_{m=1}^{M} \cup \{\mathbf{v}(E_{c})\}.
\end{equation}

The constructed graph $G_{c}$ is processed through a GCN, which updates the node embeddings. This operation can be compactly expressed as:
\begin{equation}
    H = GCN(V_{c},A_{c}),
\end{equation}
where $H$ represents the final node embeddings after two graph convolution layers, with intermediate activations and regularization (such as ReLU and dropout) applied within the GCN function.

Next, we fuse these embeddings with the emotion class embedding $\mathbf{v}(E_{c})$ using a simple attention mechanism. Specifically, the attention weights are obtained via a softmax operation, and the final emotion class embedding is obtained as a weighted average of the node embeddings:
\begin{equation}
    h(E_{c})=softmax(H \mathbf{v}(E_{c})^{T})H + H.
\end{equation}

Furthermore, to effectively capture both the overall distribution and the most salient features of each emotion class, the overall class embedding is derived by concatenating the mean and max pooling of the fused embeddings:
\begin{equation}
    F_{c} = mean(h(E_{c}))+ max(h(E_{c})).
\end{equation}

Finally, the embeddings for the positive, neutral, and negative emotion classes are concatenated to generate the final emotion semantic representation $F_{t}$.

\subsection{Visual-Semantic Interactive Module}
Given the complexity of group environments, we propose a VSIM to enhance the alignment and interaction between visual and semantic information, thereby improving recognition performance. This module incorporates two key strategies: SFF and SAM loss function. The SFF mechanism selectively emphasizes meaningful visual-semantic correlations in the final representation, while the SAM loss reinforces alignment across different emotional features by ensuring bidirectional consistency between visual and semantic information.  

\textbf{1) Similarity-aware Feature Fusion:} 
In some cases, particularly where the visual-textual relationship is weak (e.g., ambiguous or uncertain emotions), direct concatenation can introduce noise. To effectively integrate visual and semantic features, we introduce distinct projection heads, each consisting of a fully connected layer. The projected features are then used to compute the cosine similarity between the visual and semantic representations. This similarity is leveraged to modulate the intensity of the fused feature $F_{v;t}$, enhancing the integration of two branches:

\begin{equation}
\begin{aligned}
    F_{v;t} = [F_{v}; F_{t}],\\
    sim = \frac{F_{v} (F_{t})^T}{\|F_{v}\| \|F_{t}\|}.
\end{aligned}
\end{equation}

Then, we further refine the fusion process through similarity-weighted adjustment. During training, the similarity scores $sim$ are standardized by calculating their mean and standard deviation. A Sigmoid function is then applied to map the standardized scores to the range $[0, 1]$. These re-scaled similarity scores are subsequently used to reweight the fused features for enhanced context integration:

\begin{equation}
F_{v;t}^{'} = Sigmoid(Std(sim)) F_{v;t}.
\end{equation}

The adjusted feature $ F_{v;t}^{'} $ is then processed through 4-layer transformer blocks to derive the final group-level emotion representation $ F_{group} $. 

\textbf{2) Similarity Alignment Matching Loss:}
To achieve robust alignment between visual and semantic features, we introduce a similarity alignment matching loss, motivated by the bidirectional cross-modal matching method proposed in~\cite{JiangY23}. The SAM loss accounts for both visual-to-semantic and semantic-to-visual interactions.

Given a mini-batch of $ N $ samples, \ie, $ N\times N $ visual-text pairs, the similarity between each visual feature $ F_{v_i} $ and all semantic features $ \{ F_{t_j} \}_{j=1}^{N} $ is evaluated using cosine similarity:

\begin{equation}
    sim(F_{v_i}, F_{t_j}) = \frac{(F_{v_i})^{T} F_{t_j}}{\| F_{v_i} \| \| F_{t_j} \|}.
\end{equation}

Equally, the similarity between each semantic feature $ F_{t_j} $ and all visual features $ \{ F_{v_i} \}_{i=1}^{N} $ is also computed. Matching probabilities are obtained by applying a softmax function with a temperature scaling factor $ \alpha $:

\begin{equation}
    p_{i,j} = \frac{\exp(\alpha \cdot sim(F_{v_i}, F_{t_j}))}{\sum_{k=1}^{N} \exp(\alpha \cdot sim(F_{v_i}, F_{t_k}))}.
\end{equation}

The SAM loss combines both visual-to-text and text-to-visual components. For visual-to-text alignment, the loss is defined as:
\begin{equation}
    \mathcal{L}_{v \rightarrow t}  = \frac{1}{N} \sum_{i=1}^{N} \sum_{j=1}^{N} q_{i,j} \log \left(\frac{q_{i,j}}{p_{i,j} + \epsilon}\right),
\end{equation}
where $\epsilon$ is a small constant to avoid numerical instability. $ q_{i,j} $ denotes the ground truth matching probability. The matching probability is typically normalized as follows:

\begin{equation}
    q_{i,j} = \frac{y_{i,j}}{\sum_{k=1}^{N} y_{i,k}},
\end{equation}
where $ y_{i,j} $ denote the true matching label if $i$ and $j$ represent a true matching pair from the same emotion label, $ y_{i,j} = 1 $; otherwise, $ y_{i,j} = 0 $. The value $ y_{i,k} $ represents the probability that the pair is correctly matched.

Symmetrically, for text-to-visual alignment, the same operation is applied by swapping the roles of visual and semantic features. The final SAM loss is computed as:

\begin{equation}
    \mathcal{L}_{SAM} = \mathcal{L}_{v \rightarrow t} + \mathcal{L}_{t \rightarrow v}.
\end{equation}

\subsection{Training and Optimization}
Our framework adopts an end-to-end training strategy that jointly optimizes classification and emotional feature alignment. To achieve effective GER, we define two main objectives: classification loss and visual-semantic alignment loss.
The classification loss comprises four components: $\mathcal{L}_{group}$, which supervises the fused group-level feature $F_{group}$ (derived from both semantic and visual branches), and component losses $\mathcal{L}_{s}$, $\mathcal{L}_{f}$, and $\mathcal{L}_{o}$, which correspond to the classification of different visual emotion cues. Additionally, we introduce a similarity alignment matching loss, $\mathcal{L}_{SAM}$, which enforces consistent alignment between visual and semantic features. The overall optimization objective for training is defined as:
\begin{equation}
\begin{aligned}
    \mathcal{L}_{cls} = \mathcal{L}_{group} + \mathcal{L}_{s} + \mathcal{L}_{f} + \mathcal{L}_{o},\\
    \mathcal{L}_{total} = \mathcal{L}_{cls} + \mathcal{L}_{SAM},
\end{aligned}
\end{equation}
where $\mathcal{L}_{cls}$ represents the cross-entropy loss series. During inference, the model uses the refined group-level representation for emotion prediction, capturing both classification cues and cross-branch consistency to enhance GER performance.

\section{Experiments}
\subsection{Experiment Settings}
\label{4.1}
\noindent\textbf{GAFF Datasets.} The GAFF datasets comprise two benchmark datasets: Group AFFective 2.0 (GAFF2)~\citep{dhall2017individual} and Group AFFective 3.0 (GAFF3)~\citep{dhall2018emotiw}. These datasets were curated from online sources using targeted keywords such as “protest,” “violence,” and “festival” to ensure that each image contains at leasttwo individuals. Table~\ref{tab:database} summarize key statistics for each emotion category and the total number of images. Emotions are categorized as positive, negative, and neutral classes. Notably, the test set labels are withheld and accessible only to participants in the EmotiW competitions~\cite{dhall2017individual,dhall2018emotiw}. As a result, all experiments are conducted using only the training and validation sets, following the standard evaluation protocol, where models are trained on the training set and evaluated on the validation set.

\noindent\textbf{GroupEmoW Dataset.} The GroupEmoW dataset~\cite{guo2020graph} is a publicly available dataset spedifically designed for GER tasks, containing 15,894 images, divided into training, validation, and test subsets, with 11,127, 3,178, and 1,589 images in each set, respectively. These images were sourced from platforms such as Google, Baidu, Bing, and Flickr using search terms related to social events, including funerals, birthdays, protests, conferences, meetings, and weddings. Each image is annotated with one of three group-level emotions: positive, neutral, or negative. Detailed statistics for each class are presented in Table~\ref{tab:database}. Importantly, the labels for all sets are publicly available. For training and evaluation, we followed the protocols outlined in~\cite{WangZTL23, Khan0CT21}, using using the training set for model training and the test set for performance evaluation.

\noindent\textbf{Implementation Details.}
For all datasets, face regions are first detected and cropped using the MTCNN detector~\cite{zhang2016joint}, then resized to $224 \times 224$ pixels. Object proposals are generated by Faster R-CNN~\cite{RenHG017} pretrained on the MSCOCO dataset, while scene samples are uniformly resized to $256 \times 256$ pixels. The extracted features from each emotion cue and branch are standardized to a 512-dimensional vector. The hidden size and the number of heads for each layer in the multi-head attention and feed-forward network are set to 512 and 8, respectively. During training, we adopt the ADAM optimizer with hyperparameters $\beta_{1} = 0.9$, $\beta_{2} = 0.999$, and $\epsilon = 10^{-8}$, with an initial learning rate of 0.001 that decays by a factor of 0.9 per iteration. The momentum is fixed at 0.9, and the batch size is set to 4. The temperature parameter $\alpha$ in the SAM loss is set to 0.02. The entire framework is implemented in PyTorch, with training accelerated using an NVIDIA RTX 3090 GPU.

\begin{table}
\small
\centering
\caption{Statistics of the three evaluation benchmark datasets. Elements listed in the table are the number of images.}
\begin{tabular}{cccccc}		
\toprule
Dataset& Type & Pos & Neu & Neg & Total\\
\midrule
\multirow{2}*{GAFF2} & Train & 1272 & 1199 & 1159 & 3630\\
~ & Val & 773 & 728 & 564 & 2065 \\
\midrule
\multirow{2}*{GAFF3} & Train & 3977 & 3080 & 2758 & 9815\\
~ & Val & 1747 & 1368 & 1231 & 4346 \\
\midrule
\multirow{3}*{GroupEmoW} & Train & 4645 & 3463 & 3019 & 11127\\
~ & Val & 1327 & 990 & 861 & 3178 \\
~ & Test & 664 & 494 & 431 & 1589 \\

\bottomrule
\end{tabular}
\label{tab:database}
\vspace{-10pt}
\end{table}

\subsection{Comparison with State-of-the-Art Methods}
\begin{table}[t!]
\small
\centering
\caption{Comparison with the state-of-the-art methods on the GAFF2 dataset in the overall accuracy (\%). }

\begin{tabular}{cccccc}
\toprule
Source & Methods & Pos & Neu & Neg  & Overall\\
\hline
\multirow{7}{*}{Face+Scene}& Surace+ \citep{surace2017emotion} & 68.61 & 59.63 & 76.05  & 67.75\\
& Huang+ \citep{huang2019analyzing} & -- & -- & --  & 72.17\\
& Fujii+ \citep{fujii2019hierarchical} & 75.68 & 69.64 & \textbf{77.33} & 74.00\\
& Fujii+ \citep{fujii2020hierarchical} & 78.01 & 72.92 & 76.48  & 75.81\\
& Zhang+ \citep{zhang2022semi} & \textbf{85.38} & \textbf{84.49} & 60.89 & 78.51\\
& Wang+ \citep{WangZTL23} & 80.08 & 79.13 & 77.31 & 79.00\\
& Ours & 82.51 & 82.51 & 71.24 & \textbf{79.50}\\
\hline
\multirow{3}{*}{\makecell{Face+Scene\\+Object}}
& Wang+ \citep{WangZTL23} & 80.98 & \textbf{80.39} & \textbf{76.01} & 79.45\\
& Fujii+ \citep{fujii2020hierarchical} & 87.84 & 77.55& 74.10 & \textbf{80.41}\\
& Ours & \textbf{88.60} & 74.89 & 75.32 & 80.25\\
\bottomrule
\end{tabular}
\label{tab:GAFF2}
\vspace{-10pt}
\end{table}

\begin{table}[t!]
\small
\centering
\caption{The complexity comparisons with methods that exhibit relatively superior classification performance. "\# Param." indicates the total number of the training model parameters.}
\begin{tabular}{lccc}
\toprule
Method & Feature Aggregation & Classification & \# Param.\\
\midrule
Fujii+ \cite{fujii2020hierarchical} & Cascade Attention  & Binary & 356M  \\

Ours  & Transformer & Primary & 264M\\
\bottomrule
\end{tabular}
\label{tab:Complex}
\vspace{-10pt}
\end{table}

\begin{table}[t!]
\small
\centering
\caption{Comparison with the state-of-the-art methods on the GAFF3 dataset in the overall accuracy (\%).}
\begin{tabular}{cccccc}
\toprule
Source & Methods & Pos & Neu & Neg & Overall\\
\hline
\multirow{8}{*}{Face+Scene}& Fujii+ \citep{fujii2019hierarchical} & 71.12 & 69.51 & 71.52  & 71.27\\
& Quach+ \citep{quach2022non} & -- & -- & --  & 74.18\\
& Fujii+ \citep{fujii2020hierarchical} & 78.42 & 71.19 & 73.40   & 75.05\\
& Zhang+ \citep{zhang2022semi} & 79.85 & 76.61 & 73.44  & 77.01\\
& Khan+ \citep{Khan0CT21} & -- & -- & --  & 78.14\\
& Khan+ \citep{KhanLCMOT18} & 87.64 & 70.76 & 73.76  & 78.39\\
& Wang+ \citep{WangZTL23} & 86.61 & \textbf{78.36} & 69.21  & 79.08\\
& Xie+~\cite{XieLCCLSC23}  & - & - & - & 79.20\\
& Ours & \textbf{87.97} & 77.70 & \textbf{75.95}  & \textbf{81.33}\\
\hline
\multirow{6}{*}{\makecell{Face+Scene\\+Object}}& Fujii+ \citep{fujii2020hierarchical} & 82.88 & 72.64 & 74.32 & 77.54\\
& Guo+ \citep{guo2020graph} & -- & -- & --   & 78.87\\
& Guo+ \citep{GuoZPBB18} & -- & -- & -- & 78.98\\
& Khan+ \citep{Khan0CT21} & 86.89 & 75.00 & 72.71  & 79.13\\
& Wang+ \citep{WangZTL23} & 86.89 & \textbf{78.58} & 70.35 & 79.59\\
& Ours & \textbf{93.64} & 68.35 & \textbf{78.64}  & \textbf{81.42}\\
\bottomrule
\end{tabular}
\label{tab:GAFF3}
\vspace{-10pt}
\end{table}

\begin{table}[t!]
\small
\centering
\caption{Comparison with the state-of-the-art methods on the GroupEmoW dataset in the overall accuracy (\%). }
\begin{tabular}{cccccc}
\toprule
Source & Methods & Pos & Neu & Neg  & Overall\\
\hline
\multirow{4}{*}{Face+Scene}& Zhang+~\cite{zhang2022semi} & 94.13 & 85.22 & 84.22    & 88.67\\
& Khan+~\cite{Khan0CT21} & - & - & - & 89.36\\
& Wang+~\cite{WangZTL23} & 94.58 & 85.63 & 86.77 & 89.68\\
& Xie+~\cite{XieLCCLSC23}  & - & - & - & 90.47\\
& Ours & \textbf{94.72} & \textbf{87.25} & \textbf{90.02}  & \textbf{91.12}\\
\hline
\multirow{4}{*}{\makecell{Face+Scene\\+Object}}
& Guo+ \citep{guo2020graph} & -- & -- & -- & 89.93\\
& Wang+ \citep{WangZTL23} & 94.58 & 85.46 & 88.63 & 90.06\\
& Khan+ \citep{Khan0CT21}& -- & -- & -- &  90.18\\
& Ours & \textbf{96.68} & \textbf{86.03} & \textbf{88.63} & \textbf{91.18} \\
\bottomrule
\end{tabular}
\label{tab:GroupEmoW}
\vspace{-10pt}
\end{table}

We compared the proposed framework with several state-of-the-art GER approaches, reflecting the latest advancements in this field. These methods can be grouped into two categories: (1) independently learning emotion representations from different visual emotion cues \cite{ quach2022non, huang2019analyzing, surace2017emotion, fujii2019hierarchical,  KhanLCMOT18, guo2020graph}; and (2) enhancing individual representations by encoding global scene contextual information using different techniques, such as GNNs \cite{guo2020graph} or attention mechanisms \cite{fujii2020hierarchical, zhang2022semi, WangZTL23,  Khan0CT21, GuoZPBB18, XieLCCLSC23}. To ensure a fair comparison, we assess performance under consistent conditions in the visual branch, which can be broadly classified into two types: (1) methods that integrate face and scene cues, and (2) methods that integrate face, scene, and object cues.

\textbf{1) Results on the GAFF2 dataset:} The performance comparisons of various methods are summarized in Table~\ref{tab:GAFF2}. In the scenario where the visual branch uses both face and scene cues, our proposed method achieves 79.50\% on the overall result, surpassing state-of-the-art methods on the GAFF2 dataset. However, the recognition accuracy of the proposed method for the positive and neutral categories is lower than that of the Zhang~\etal~\cite{zhang2022semi}. This is can be attributed to the fact that Zhang~\etal~\cite{zhang2022semi} adopts a contrastive learning strategy, which not only extracts effective features from labeled images but also leverages information from unlabeled images, significantly boosting the model's ability to recognize emotions in the positive and neutral categories. Nevertheless, the recognition accuracy of the negative category is higher in our proposed method compared to Zhang~\etal~\cite{zhang2022semi}. This may be because Zhang~\etal~\cite{zhang2022semi} struggles to effectively address the confusion between neutral and negative categories. In contrast, our approach enhances recognition by integrating global contextual cues with individual features while leveraging label-driven semantics to refine emotion embeddings, effectively boosting GER performance. 

Furthermore, when compared to the results obtained using all three emotion cues (face, object, and scene) in the visual branch, our method achieves competitive performance, recording an overall result of 80.25\%. Although our approach slightly underperforms the hierarchical classification strategy in \cite{fujii2020hierarchical}, it is worth noting that even the top-performing method by Wang~\etal~\cite{WangZTL23} does not surpass \cite{fujii2020hierarchical} under similar conditions. Fujii~\etal initially conducted binary classification using aggregated facial features and then extended their approach to a three-class classification, incorporating object and scene features while harnessing computational resources to capture intricate object details. In contrast, our method utilizes a lightweight CNN-based encoder for direct three-class classification of object information, optimizing computational efficiency without sacrificing performance. Additionally, Table~\ref{tab:Complex} presents a comparative analysis of the method complexity between our approach and that of Fujii~\etal~\cite{fujii2020hierarchical}, demonstrating that our method achieves competitive recognition performance while significantly reducing computational complexity. Moreover, our method notably improves accuracy in positive category, highlighting its efficacy in capturing positive emotion via using LLMs with emotion labels to generate fine-grained emotion lexicons. However, in neutral category, where emotional cues are less pronounced, this approach may cause the model to deviate since it struggles to fully capture their characteristics, leading to a noticeable performance drop in these categories. 

\textbf{2) Results on the GAFF3 dataset:} Table~\ref{tab:GAFF3} presents a comparison with various methods on the GAFF3 dataset. Our method outperforms the state-of-the-art methods, achieving the highest overall result. Specifically, when comparing the performance of methods that utilize face and scene cues in the visual branch, our model demonstrates significant improvements, reaching the highest overall result of 81.33\%. This outperforms two current notably proficient methods, Wang~\etal~\cite{WangZTL23} and Xie~\etal~\cite{XieLCCLSC23}, which yielded overall accuracies of 79.08\% and 79.20\%, respectively. However, the proposed method performs relatively poorly in identifying neutral category. On the other hand, it achieves a significant performance boost in both the positive and negative categories. This can be attributed to the method's ability to enrich emotional semantic information using LLMs to generate fine-grained emotion lexicons. These lexicons offer higher discrimination in expressing strong emotions, thus enhancing the model's ability to recognize positive and negative emotions. Nevertheless, the model may struggle with words that have ambiguous or neutral emotional connotations, leading to a decline in recognition accuracy for the neutral category. Similarly, when enhanced with the additional object cue in the visual branch, our method maintains competitive performance across all metrics, achieving an overall accuracy of 81.42\%, surpassing the closest competitor by 1.83\%.  

\textbf{3) Results on the GroupEmoW dataset:} Table~\ref{tab:GroupEmoW} presents a detailed comparison of our approach with several state-of-the-art methods on the GroupEmoW dataset. When utilizing only face and scene features, our model achieves notable performance across all emotion categories, with a particularly significant improvement in the neutral (87.25\%) and negative (90.02\%) classes, surpassing the best-performing prior method by 1.62\% and 3.25\%, respectively. This leads to an overall accuracy increase of 1.44\%, indicating better overall class balance. Incorporating object features alongside face and scene cues results in further performance enhancements. Specifically, in the positive class, our method achieves a new high of 96.68\%, outperforming all previous methods. Notably, the neutral class improves to 86.03\%, while the negative class remains competitive at 88.63\%. As a result, our method achieves the highest overall accuracy of 91.18\%. Collectively, these results highlight the effectiveness of our method in improving GER performance. They demonstrate the robustness and generalization capability of our model in handling complex emotional scenarios that involve multiple emotion cues and branches. This improvements can be attributes to the integration of comprehensive scene context and label-related semantics, which together enhance the model's ability to interpret nuanced emotional signals.

\begin{table*}[]
\centering
\caption{Ablation results on the GAFF2 dataset by using
different modules.}
\begin{tabular}{clccccccc}
\toprule
\multirow{2}{*}{No.} & \multirow{2}{*}{Methods} & \multicolumn{3}{c}{Module} & \multirow{2}{*}{Pos} & \multirow{2}{*}{Neu} & \multirow{2}{*}{Neg}  & \multirow{2}{*}{Overall}\\
\cline{3-5}
 & & VCEM & ESEM & VSIM & & & &\\
\hline
1 & B1 &  &   &  & 82.25 & 58.25 & 84.42  & 74.41\\
2 & B2 w/o CAM & \checkmark &  &  & 81.22 & 75.60 & 75.32  & 77.67\\
3 & B2 & \checkmark & & & 78.37 & 72.92 & 86.64  & 78.66\\
4 & B3 & \checkmark & \checkmark & & 84.72 & 64.60 & 85.34  & 77.82\\
5 & B4 w/o $L_{SAM}$ & \checkmark & \checkmark & \checkmark & 79.40 & 79.83 & 76.62  & 78.81\\
6 & B4 w/o SFF & \checkmark & \checkmark & \checkmark & 81.61 & 74.89 & 82.56  & 79.50\\
7 & B4 (ours) & \checkmark & \checkmark & \checkmark & 88.60 & 74.89 & 75.32 & 80.25\\
\bottomrule
\end{tabular}
\label{tab:abaltion}
\vspace{-10pt}
\end{table*}

\section{Ablation Studies}
\textbf{1) Effects of the Proposed Modules:} To better understand the contribution of each module in our method, we conduct the following ablation studies.
\begin{itemize}
    \item B1 (Base model): This base model includes backbones corresponding to the face, object, and scene cues, along with a final softmax classification layer. This is the only visual branch.
    \item B2 (B1+VCEM): This method integrates the B1 with VCEM. This variant integrates the CAM to enhance individual representations by leveraging multi-scale context information. Preceding the final softmax classification layer, both emotion cues in the visual branch undergo interaction through a standard transformer encoder.
    \item B3 (B2+ESEM): This method consists of the network B2, and ESEM. This variant, while employing LLM, it generates fine-grained emotion lexicons and models their associations with the emotion labels, thereby enhancing the semantic information.
    \item B4 (Ours): The method encompasses the network B3, complemented by the VSIM. Within VSIM, both the visual and the semantic information are incorporated to gradually obtain enhanced group-level emotion representation.
\end{itemize}

The results in Table~\ref{tab:abaltion} demonstrate the effectiveness of each core module through comparisons across key model variants. Specifically, integrating the VCEM into the baseline (“No. 3” vs. “No. 1”) leads to a significant improvement in overall performance, increasing from 74.41\% to 78.66\%. This demonstrates the importance of capturing rich visual contextual information in GER. However, incorporating the ESEM (“No. 4” vs. “No. 3”) results in a slight performance drop of 0.84\%. This decline can be attributed to the direct concatenation of features from the visual and semantic branches, which lacks proper alignment and fusion of the two modalities, potentially introducing noise that offsets the benefits of lexicon-based semantic representations generated by LLM. The most significant improvement occurs with the introduction of VSIM (No. 7” vs. “No. 4), yielding a gain of 2.43\% and achieving an overall accuracy of 80.25\%. These results emphasize that integrating both visual scene context and label-based semantic features, alongside a similarity-aware alignment and fusion strategy, significantly enhances  group-level emotion representation, leading to notable improvements in GER performance.

\textbf{2) Variants of Semantic and Visual Information Fusion:} To evaluate the effectiveness of our VSIM, we conducted ablation studies on different fusion strategies, as shown in experiments “No. 5” to “No. 7” in Table~\ref{tab:abaltion}. The comparison between “No. 4” and “No. 5” indicates the importance of adopting a similarity-weighted adjustment mechanism rather than straightforward concatenation for better integration of visual and semantic information. Further enhancing this alignment, experiment “No. 6” incorporates the $L_{SAM}$ loss, resulting in a notable improvement in recognition accuracy, achieving an overall performance of 79.50\%. By combining both SFF and $L_{SAM}$ in “No. 7”, the model demonstrates a stronger ability to capture complex interactions, leading to a 2.43\% increase in overall performance compared to “No. 4”.

\textbf{3) Effect of Visual Context Encoding and Semantic Embedding Variants:} The results in Table~\ref{tab:abaltion} illustrate the performance improvements achieved by incorporating visual context encoding and semantic embeddings. Specifically, for the VCEM, the comparison between “No. 3” and “No. 2” shows that integrating the CAM mechanism increases the overall performance from 77.67\% to 78.66\%. This enhancement highlights the effectiveness of encoding multi-scale visual context information, enabling richer and more nuanced emotion representations compared to using only a standard transformer encoder for fusing different emotion cues.

\begin{figure}
    \centering
    \includegraphics[width=1.0\linewidth]{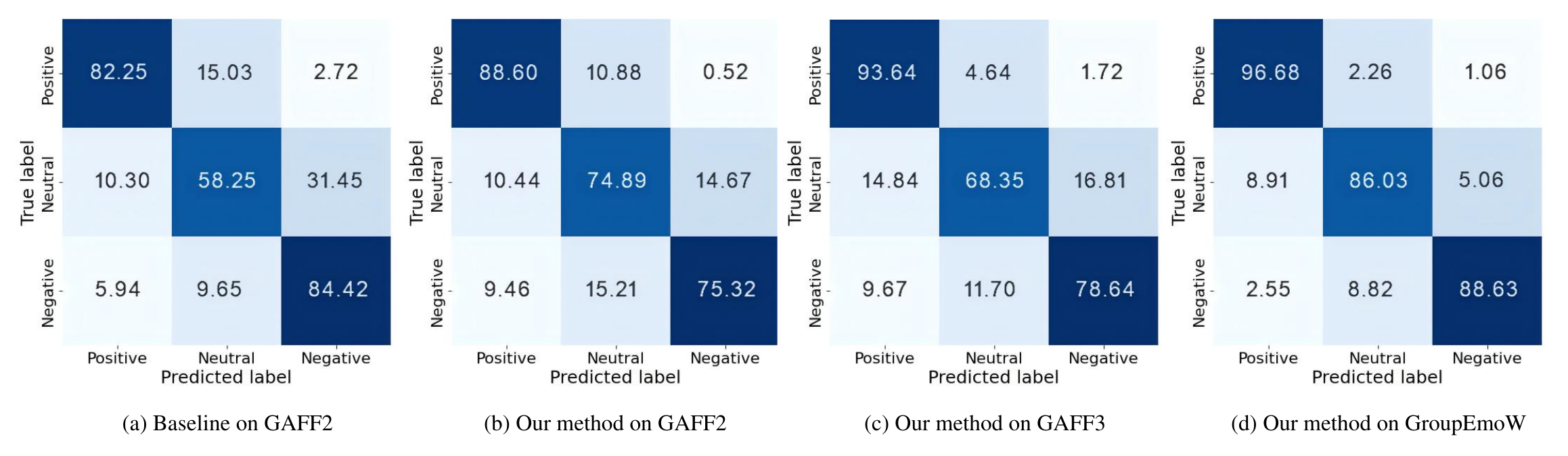}
    \caption{Comparison of the confusion matrices on the GAFF2, GAFF3, and GroupEmoW datasets.}
    \label{fig:confusion_matrices}
    \vspace{-10pt}
\end{figure}

\begin{figure}
    \centering
    \includegraphics[width=1.0\linewidth]{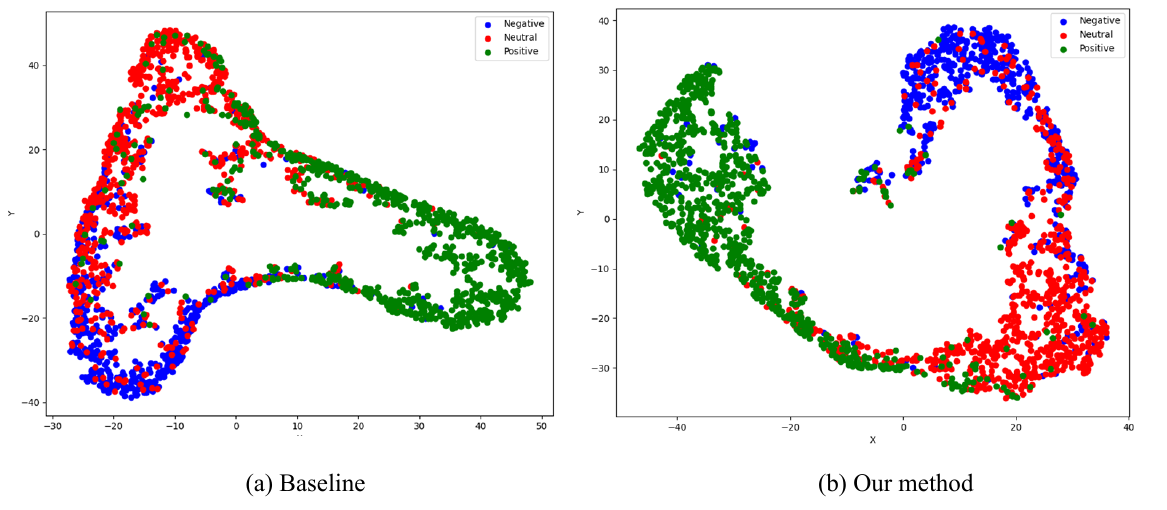}
    \caption{The t-SNE visualization of learned representation by different models on the GAFF2 dataset (color viewed is best).}
    \label{fig:feat_vis}
    \vspace{-10pt}
\end{figure}

\section{Visualization and Analysis}
\textbf{1) The visualization of confusion matrices.} Fig.~\ref{fig:confusion_matrices} illustrates a comparative analysis of classification performance between the baseline and our proposed methods on the GAFF2, GAFF3, and GroupEmoW datasets, focusing on three emotion categories: negative, neutral, and positive. As shown in Fig.~\ref{fig:confusion_matrices}(a), the baseline model on the GAFF2 dataset achieves high accuracy for the negative class (84.42\%) but struggles to differentiate between negative and neutral emotions, resulting in a high misclassification rate of31.45\%. While the positive class attains a reasonable accuracy of 82.25\%, the neutral class remains challenging, with an accuracy of only 58.25\%. In contrast, Fig.~\ref{fig:confusion_matrices}(b) demonstrates that our proposed method on the GAFF2 dataset significantly mitigates this confusion. While maintaining a strong recognition rate for the negative class (75.32\%), it is notably enhances the  neutral class accuracy, increasing it to 74.89\%.  Fig.~\ref{fig:confusion_matrices}(c) highlights that the effectiveness of our method on GAFF3, achieving high accuracies for both negative (78.64\%) and positive (93.64\%) classes, demonstrating improved differentiation between these categories. Similarly, in Fig.~\ref{fig:confusion_matrices}(d), our proposed method on the GroupEmoW dataset achieves 96.68\% accuracy for the positive class, with minimal confusion between categories. Additionally, the neutral and negative classes attain accuracies of 86.03\% and 88.63\%, respectively, effectively reducing misclassification compared to the GAFF2 and GAFF3 datasets. 

Overall, these results demonstrate the robustness of our method across multiple datasets, consistently improving classification accuracy, particularly by reducing confusion between similar emotional states.

\textbf{2) The t-SNE visualization of learned representation.} 
The t-SNE~\cite{van2008visualizing} visualizations in Fig.~\ref{fig:feat_vis} compare the feature distribution learned by the baseline model (a) and our proposed method (b) for GER on the GAFF2 dataset. In the baseline model, the features representing the three emotion categories (positive, neutral, and negative) overlap significantly, indicating poor class separation and weak discriminative power. In contrast, our method demonstrates a clear and distinct separation between the feature clusters, especially enhancing the differentiation between neutral and positive emotions. This improved feature segregation highlights the effectiveness of our approach in capturing and encoding relevant emotional cues, ultimately boosting classification performance. 

\section{Conclusion}
In this paper, we propose a novel framework for group-level emotion recognition that effectively integrates multi-scale visual scene context and label-guided semantic information. Our approach introduces two key modules: the Visual Context Encoding Module (VCEM) and Emotion Semantic Encoding Module (ESEM), addressing the critical limitations of existing GER methods, which often underexploit contextual and semantic cues essential for accurate emotion inference. The VCEM enhances individual features by incorporating multi-scale global context, while the ESEM module refines semantic representations through label-guided emotion lexicons generated by a LLM. To further bridge the gap between visual and semantic representations, we propose the Visual-Semantic Interaction Module (VSIM), which aligns and fuses these features to produce a comprehensive group-level emotion representation. Extensive experiments on three public benchmarks demonstrate the effectiveness of our approach, yielding significant performance improvements over state-of-the-art methods. For future work, we will explore extending our framework to more complex and diverse real-world scenarios and exploring advanced multi-modal feature fusion techniques to further enhance GER performance.


\footnotesize
\bibliographystyle{IEEEtran}
\normalem
\bibliography{IEEEtemplate}

\begin{thebibliography}{10}
\providecommand{\url}[1]{#1}
\csname url@samestyle\endcsname
\providecommand{\newblock}{\relax}
\providecommand{\bibinfo}[2]{#2}
\providecommand{\BIBentrySTDinterwordspacing}{\spaceskip=0pt\relax}
\providecommand{\BIBentryALTinterwordstretchfactor}{4}
\providecommand{\BIBentryALTinterwordspacing}{\spaceskip=\fontdimen2\font plus
\BIBentryALTinterwordstretchfactor\fontdimen3\font minus
  \fontdimen4\font\relax}
\providecommand{\BIBforeignlanguage}[2]{{%
\expandafter\ifx\csname l@#1\endcsname\relax
\typeout{** WARNING: IEEEtran.bst: No hyphenation pattern has been}%
\typeout{** loaded for the language `#1'. Using the pattern for}%
\typeout{** the default language instead.}%
\else
\language=\csname l@#1\endcsname
\fi
#2}}
\providecommand{\BIBdecl}{\relax}
\BIBdecl

\bibitem{TurchetOOG24}
L.~Turchet, B.~O'Sullivan, R.~Ortner, and C.~Guger, ``Emotion recognition of
  playing musicians from eeg, ecg, and acoustic signals,'' \emph{{IEEE} Trans.
  Hum. Mach. Syst.}, vol.~54, no.~5, pp. 619--629, 2024.

\bibitem{SongGYTW24}
P.~Song, D.~Guo, X.~Yang, S.~Tang, and M.~Wang, ``Emotional video captioning
  with vision-based emotion interpretation network,'' \emph{{IEEE} Trans. Image
  Process.}, vol.~33, pp. 1122--1135, 2024.

\bibitem{johnson2021exploring}
D.~Johnson, S.~Thurairajasingam, V.~Letchumanan, K.-G. Chan, and L.-H. Lee,
  ``Exploring the role and potential of probiotics in the field of mental
  health: major depressive disorder,'' \emph{Nutrients}, vol.~13, no.~5, p.
  1728, 2021.

\bibitem{YinJHYW24}
Y.~Yin, L.~Jing, F.~Huang, G.~Yang, and Z.~Wang, ``{MSA-GCN:} multiscale
  adaptive graph convolution network for gait emotion recognition,''
  \emph{Pattern Recognit.}, vol. 147, p. 110117, 2024.

\bibitem{RodriguezFP25}
L.~A.~O. Rodr{\'{\i}}guez, R.~G. Fern{\'{a}}ndez, and D.~M. Palacio,
  ``Individual performance in women's grassroots football: {A} physical and
  emotional perspective,'' \emph{{IEEE} Trans. Hum. Mach. Syst.}, vol.~55,
  no.~1, pp. 83--92, 2025.

\bibitem{MaoZZSH22}
Q.~Mao, L.~Zhou, W.~Zheng, X.~Shao, and X.~Huang, ``Objective class-based
  micro-expression recognition under partial occlusion via region-inspired
  relation reasoning network,'' \emph{{IEEE} Trans. Affect. Comput.}, vol.~13,
  no.~4, pp. 1998--2016, 2022.

\bibitem{barsade2015group}
S.~G. Barsade and A.~P. Knight, ``Group affect,'' \emph{Annu. Rev. Organ.
  Psychol. Organ. Behav.}, vol.~2, no.~1, pp. 21--46, 2015.

\bibitem{barsade2012group}
S.~G. Barsade and D.~E. Gibson, ``Group affect: Its influence on individual and
  group outcomes,'' \emph{Curr. Dir. Psychol.}, vol.~21, no.~2, pp. 119--123,
  2012.

\bibitem{SanchezHTH20}
F.~L. S{\'{a}}nchez, I.~Hupont, S.~Tabik, and F.~Herrera, ``Revisiting crowd
  behaviour analysis through deep learning: Taxonomy, anomaly detection, crowd
  emotions, datasets, opportunities and prospects,'' \emph{Inf. Fusion},
  vol.~64, pp. 318--335, 2020.

\bibitem{VeltmeijerGH23}
E.~Veltmeijer, C.~Gerritsen, and K.~V. Hindriks, ``Automatic emotion
  recognition for groups: {A} review,'' \emph{{IEEE} Trans. Affect. Comput.},
  vol.~14, no.~1, pp. 89--107, 2023.

\bibitem{huang2019analyzing}
X.~Huang, A.~Dhall, R.~Goecke, M.~Pietik{\"a}inen, and G.~Zhao, ``Analyzing
  group-level emotion with global alignment kernel based approach,''
  \emph{{IEEE} Trans. Affect. Comput.}, vol.~13, no.~2, pp. 713--728, 2019.

\bibitem{KhanLCMOT18}
A.~Khan, Z.~Li, J.~Cai, Z.~Meng, J.~O'Reilly, and Y.~Tong, ``Group-level
  emotion recognition using deep models with {A} four-stream hybrid network,''
  in \emph{Proc. {ACM} ICMI}, 2018, pp. 623--629.

\bibitem{surace2017emotion}
L.~Surace, M.~Patacchiola, E.~Battini~S{\"o}nmez, W.~Spataro, and A.~Cangelosi,
  ``Emotion recognition in the wild using deep neural networks and bayesian
  classifiers,'' in \emph{Proc. {ACM} ICMI}, 2017, pp. 593--597.

\bibitem{fujii2020hierarchical}
K.~Fujii, D.~Sugimura, and T.~Hamamoto, ``Hierarchical group-level emotion
  recognition,'' \emph{{IEEE} Trans. Multi.}, vol.~23, pp. 3892--3906, 2020.

\bibitem{Khan0CT21}
A.~S. Khan, Z.~Li, J.~Cai, and Y.~Tong, ``Regional attention networks with
  context-aware fusion for group emotion recognition,'' in \emph{Proc. WACV},
  2021, pp. 1150--1159.

\bibitem{zhang2022semi}
J.~Zhang, X.~Wang, D.~Zhang, and D.-J. Lee, ``Semi-supervised group emotion
  recognition based on contrastive learning,'' \emph{Electronics}, vol.~11,
  no.~23, p. 3990, 2022.

\bibitem{WangZTL23}
X.~Wang, D.~Zhang, H.~Tan, and D.~Lee, ``A self-fusion network based on
  contrastive learning for group emotion recognition,'' \emph{{IEEE} Trans.
  Comput. Soc. Syst.}, vol.~10, no.~2, pp. 458--469, 2023.

\bibitem{guo2020graph}
X.~Guo, L.~Polania, B.~Zhu, C.~Boncelet, and K.~Barner, ``Graph neural networks
  for image understanding based on multiple cues: Group emotion recognition and
  event recognition as use cases,'' in \emph{Proc. WACV}, 2020, pp. 2921--2930.

\bibitem{GuoZPBB18}
X.~Guo, B.~Zhu, L.~F. Polan{\'{\i}}a, C.~Boncelet, and K.~E. Barner,
  ``Group-level emotion recognition using hybrid deep models based on faces,
  scenes, skeletons and visual attentions,'' in \emph{Proc. {ACM} ICMI}, 2018,
  pp. 635--639.

\bibitem{quach2022non}
K.~G. Quach, N.~Le, C.~N. Duong, I.~Jalata, K.~Roy, and K.~Luu, ``Non-volume
  preserving-based fusion to group-level emotion recognition on crowd videos,''
  \emph{Pattern Recognit.}, vol. 128, p. 108646, 2022.

\bibitem{DaiLSL19}
Y.~Dai, X.~Liu, D.~Shuzhan, and L.~Yang, ``Group emotion recognition based on
  global and local features,'' \emph{{IEEE} Access}, vol.~7, pp.
  111\,617--111\,624, 2019.

\bibitem{XieLCCLSC23}
H.~Xie, M.~Lee, T.~Chen, H.~Chen, H.~Liu, H.~Shuai, and W.~Cheng, ``Most
  important person-guided dual-branch cross-patch attention for group affect
  recognition,'' in \emph{Proc. ICCV}.\hskip 1em plus 0.5em minus 0.4em\relax
  {IEEE}, 2023, pp. 20\,541--20\,551.

\bibitem{ZhangCSW22}
Y.~Zhang, M.~Chen, J.~Shen, and C.~Wang, ``Tailor versatile multi-modal
  learning for multi-label emotion recognition,'' in \emph{Proc. {AAAI}
  Conference on Artificial Intelligence}, 2022, pp. 9100--9108.

\bibitem{JiangLZL23}
W.~Jiang, X.~Liu, W.~Zheng, and B.~Lu, ``Multimodal adaptive emotion
  transformer with flexible modality inputs on {A} novel dataset with
  continuous labels,'' in \emph{Proc. {ACM} International Conference on
  Multimedia,}, 2023, pp. 5975--5984.

\bibitem{LiNZ023}
H.~Li, H.~Niu, Z.~Zhu, and F.~Zhao, ``Intensity-aware loss for dynamic facial
  expression recognition in the wild,'' in \emph{Proc. {AAAI} Conference on
  Artificial Intelligence}, 2023, pp. 67--75.

\bibitem{ChenHHK23}
C.~Chen, T.~Hung, Y.~Hsu, and L.~Ku, ``Label-aware hyperbolic embeddings for
  fine-grained emotion classification,'' in \emph{Proc. Annual Meeting of the
  Association for Computational Linguistics}, 2023, pp. 10\,947--10\,958.

\bibitem{DengR23a}
J.~Deng and F.~Ren, ``Multi-label emotion detection via emotion-specified
  feature extraction and emotion correlation learning,'' \emph{{IEEE} Trans.
  Affect. Comput.}, vol.~14, no.~1, pp. 475--486, 2023.

\bibitem{PengWKNLC22}
Y.~Peng, W.~Wang, W.~Kong, F.~Nie, B.~Lu, and A.~Cichocki, ``Joint feature
  adaptation and graph adaptive label propagation for cross-subject emotion
  recognition from {EEG} signals,'' \emph{{IEEE} Trans. Affect. Comput.},
  vol.~13, no.~4, pp. 1941--1958, 2022.

\bibitem{WangLSLTF22}
K.~Wang, Z.~Lian, L.~Sun, B.~Liu, J.~Tao, and Y.~Fan, ``Emotional reaction
  analysis based on multi-label graph convolutional networks and dynamic facial
  expression recognition transformer,'' in \emph{Proc. International on
  Multimodal Sentiment Analysis Workshop and Challenge}.\hskip 1em plus 0.5em
  minus 0.4em\relax {ACM}, 2022, pp. 75--80.

\bibitem{SongGYTYW23}
P.~Song, D.~Guo, X.~Yang, S.~Tang, E.~Yang, and M.~Wang, ``Emotion-prior
  awareness network for emotional video captioning,'' in \emph{Proc. {ACM}
  {MM}}, 2023, pp. 589--600.

\bibitem{ChenGZSQDL22}
J.~Chen, Y.~Guo, J.~Zhu, G.~Sun, D.~Qin, M.~Deng, and H.~Liu, ``Improving
  few-shot remote sensing scene classification with class name semantics,''
  \emph{{IEEE} Trans. Geosci. Remote. Sens.}, vol.~60, pp. 1--12, 2022.

\bibitem{LiuWJCL24}
Y.~Liu, X.~Wang, B.~Jiang, L.~Chen, and B.~Luo, ``Semanticformer: Hyperspectral
  image classification via semantic transformer,'' \emph{Pattern Recognit.
  Lett.}, vol. 179, pp. 1--8, 2024.

\bibitem{PengYXWX24}
F.~Peng, X.~Yang, L.~Xiao, Y.~Wang, and C.~Xu, ``Sgva-clip: Semantic-guided
  visual adapting of vision-language models for few-shot image
  classification,'' \emph{{IEEE} Trans. Multim.}, vol.~26, pp. 3469--3480,
  2024.

\bibitem{ZhaoP23}
Z.~Zhao and I.~Patras, ``Prompting visual-language models for dynamic facial
  expression recognition,'' in \emph{Proc. British Machine Vision Conference},
  2023, p.~98.

\bibitem{LiuZ021}
T.~Liu, R.~Zhao, and K.~Lam, ``Multimodal-semantic context-aware graph neural
  network for group activity recognition,'' in \emph{Proc. {IEEE} ICME}.\hskip
  1em plus 0.5em minus 0.4em\relax {IEEE}, 2021, pp. 1--6.

\bibitem{WuTXGS24}
L.~Wu, M.~Tian, Y.~Xiang, K.~Gu, and G.~Shi, ``Learning label semantics for
  weakly supervised group activity recognition,'' \emph{{IEEE} Trans. Multim.},
  vol.~26, pp. 6386--6397, 2024.

\bibitem{ChengWTZ25}
K.~Cheng, L.~Wei, J.~Tang, and Y.~Zhan, ``Constraint embedding for prompt
  tuning in vision-language pre-trained model,'' \emph{Multim. Syst.}, vol.~31,
  no.~1, p.~11, 2025.

\bibitem{JungSCR024}
J.~Jung, R.~S. Sharma, W.~Chen, B.~Raj, and S.~Watanabe, ``Augsumm: Towards
  generalizable speech summarization using synthetic labels from large language
  models,'' in \emph{Proc. {IEEE} ICASSP}.\hskip 1em plus 0.5em minus
  0.4em\relax {IEEE}, 2024, pp. 12\,071--12\,075.

\bibitem{zhang2016joint}
K.~Zhang, Z.~Zhang, Z.~Li, and Y.~Qiao, ``Joint face detection and alignment
  using multitask cascaded convolutional networks,'' \emph{IEEE Signal Process.
  Lett.}, vol.~23, no.~10, pp. 1499--1503, 2016.

\bibitem{RenHG017}
S.~Ren, K.~He, R.~B. Girshick, and J.~Sun, ``Faster {R-CNN:} towards real-time
  object detection with region proposal networks,'' \emph{{IEEE} Trans. Pattern
  Anal. Mach. Intell.}, vol.~39, no.~6, pp. 1137--1149, 2017.

\bibitem{LinDGHHB17}
T.~Lin, P.~Doll{\'{a}}r, R.~B. Girshick, K.~He, B.~Hariharan, and S.~J.
  Belongie, ``Feature pyramid networks for object detection,'' in \emph{Proc.
  CVPR}, 2017, pp. 936--944.

\bibitem{HeGDG17}
K.~He, G.~Gkioxari, P.~Doll{\'{a}}r, and R.~B. Girshick, ``Mask {R-CNN},'' in
  \emph{Proc. ICCV}, 2017, pp. 2980--2988.

\bibitem{PenningtonSM14}
J.~Pennington, R.~Socher, and C.~D. Manning, ``Glove: Global vectors for word
  representation,'' in \emph{Proc. Conference on Empirical Methods in Natural
  Language Processing}, 2014, pp. 1532--1543.

\bibitem{JiangY23}
D.~Jiang and M.~Ye, ``Cross-modal implicit relation reasoning and aligning for
  text-to-image person retrieval,'' in \emph{Proc. CVPR}.\hskip 1em plus 0.5em
  minus 0.4em\relax {IEEE}, 2023, pp. 2787--2797.

\bibitem{dhall2017individual}
A.~Dhall, R.~Goecke, S.~Ghosh, J.~Joshi, J.~Hoey, and T.~Gedeon, ``From
  individual to group-level emotion recognition: Emotiw 5.0,'' in \emph{Proc.
  {ACM} ICMI}, 2017, pp. 524--528.

\bibitem{dhall2018emotiw}
A.~Dhall, A.~Kaur, R.~Goecke, and T.~Gedeon, ``Emotiw 2018: Audio-video,
  student engagement and group-level affect prediction,'' in \emph{Proc. {ACM}
  ICMI}, 2018, pp. 653--656.

\bibitem{fujii2019hierarchical}
K.~Fujii, D.~Sugimura, and T.~Hamamoto, ``Hierarchical group-level emotion
  recognition in the wild,'' in \emph{Proc. {IEEE} International Conference on
  Automatic Face {\&} Gesture Recognition}, 2019, pp. 1--5.

\bibitem{van2008visualizing}
L.~Van~der Maaten and G.~Hinton, ``Visualizing data using t-sne.''
  \emph{Journal of machine learning research}, vol.~9, no.~11, 2008.

\end{thebibliography}

\end{document}